%% file: main.tex
\documentclass[letterpaper, 10pt, conference]{ieeeconf}      

\IEEEoverridecommandlockouts                              

\usepackage{booktabs}       
\usepackage{amsfonts}       
\usepackage{multicol}
\usepackage{multirow}
\usepackage{xcolor}         
\usepackage{xspace}
\usepackage{adjustbox}
\usepackage{tabularx}
\usepackage{caption}
\usepackage{subcaption}
\usepackage{graphicx}
\usepackage{placeins}
\usepackage[absolute,showboxes]{textpos} 

\TPMargin*{3pt}
\newcommand\copyrighttext{
    \footnotesize
    \noindent
    (c) IEEE 2024\\
    Personal use of this material is permitted.
    Permission must be obtained for all other uses, in any current or future media, including reprinting/republishing this material for advertising or promotional purposes, creating new collective works, for resale or redistribution to servers or lists, or reuse of any copyrighted component of this work in other works.}%
\newcommand\copyrightnotice{%
    \begin{textblock*}{7in}(0.75in,0.15in)
        \copyrighttext
    \end{textblock*}
}

\title{\LARGE \bf Stay on Track: A Frenet Wrapper to Overcome Off-road Trajectories in Vehicle Motion Prediction
}

\author{Marcel Hallgarten$^{1,2}$ Ismail Kisa$^{2}$ Martin Stoll$^{1}$ Andreas Zell$^{2}$
\thanks{$^{1}$ Robert Bosch GmbH, Corporate Research, Renningen, Germany {\tt\small first-name.last-name@de.bosch.com}}%
\thanks{$^{2}$ University of Tuebingen, Germany {\tt\small first-name.last-name@uni-tuebingen.de}}%
}

\newcommand{\multipath}{Multipath++~\cite{varadarajan2022multipath++}\xspace}
\newcommand{\lanegcn}{LaneGCN~\cite{liang2020learning}\xspace}

\begin{document}

\maketitle
\copyrightnotice

\begin{abstract}
Predicting the future motion of surrounding vehicles is a crucial enabler for safe autonomous driving.
The field of motion prediction has seen large progress recently with State-of-the-Art (SotA) models achieving impressive results on large-scale public benchmarks.
However, recent work revealed that learning-based methods are prone to predict off-road trajectories in challenging scenarios.
These can be created by perturbing existing scenarios with additional turns in front of the target vehicle while the motion history is left unchanged.
We argue that this indicates that SotA models do not consider the map information sufficiently and demonstrate how this can be solved by representing the model inputs and outputs in a Frenet frame defined by lane centreline sequences.
To this end, we present a general wrapper that leverages a Frenet representation of the scene, and that can be applied to SotA models without changing their architecture.
We demonstrate the effectiveness of this approach in a comprehensive benchmark comprising two SotA motion prediction models.
Our experiments show that this reduces the off-road rate in challenging scenarios by more than 90\%,
without sacrificing average performance.
Code and supplementary material are available under: {\color{magenta} \texttt{https://mh0797.github.io/stayontrack/}}.
\end{abstract}

\input{chapters/1_introduction}
\input{chapters/2_related_work}
\input{chapters/3_notation}
\input{chapters/4_method}
\input{chapters/5_evaluation}
\input{chapters/6_results}
\input{chapters/7_conclusion}
\input{chapters/8_limitations}
\bibliographystyle{IEEEtran}
\bibliography{IEEEabrv,bibliography}
\input{chapters/appendix.tex}
\end{document}

%% file: chapters/1_introduction.tex
\section{Introduction}
\subsection{Generalisation Problem in the Trajectory Prediction Task} 
Self-driving vehicles need the ability to predict the future motion of other road users in order to progress collision-freely, fast, and comfortably towards their goal.
This makes the prediction of the future behaviour of surrounding road users a fundamental enabler for safe autonomous driving.
Commonly high-definition map information and past observations are used to approach this task~\cite{hagedorn2023rethinking}. 
While simple heuristics, e.g.\ constant velocity forecasting, are lightweight and fast methods, they are unable to model interactions between traffic participants and often give inaccurate predictions.
In contrast, state-of-the-art (SotA) data-driven methods~\cite{gao2020vectornet, gilles2022gohome, deo2022multimodal, liang2020learning, janjovs2023conditional} achieve excellent performance on large-scale public benchmarks~\cite{chang2019argoverse, caesar2020nuscenes}.
Various methods exist to generate critical driving scenarios based on adversarial agents~\cite{abeysirigoonawardena2019generating,wang2021advsim}, or difficult initialisations~\cite{hallgarten2024can,althoff2018automatic,klischat2019generating}.
Recent work specifically targets trajectory prediction algorithms.
In \cite{zhang2022adversarial}, an adversarial attack on trajectory prediction based on perturbations of the neighbouring vehicle's trajectory is proposed, resulting in dramatically increased prediction errors.
Similarly,~\cite{bahari2022vehicle} shows that off-road predictions increase when the road ahead is perturbed with a curve, while the motion history and the road behind are left unchanged.
This indicates that models strongly rely on extrapolating the motion history, which is reasonable in most scenarios (middle of corner, straight driving) but causes failures when a turn is ahead and the vehicle has not yet started to turn in.
The potentially misleading correlation between observed motion history and future trajectory is a well-known problem in robot behaviour prediction~\cite{muller2005off,codevilla2019exploring}.

\subsection{Mitigation Techniques}
One straightforward way to address this problem is to balance the training dataset so that challenging scenarios appear more frequently during training. As this would result in filtering out large parts of the learning dataset (e.g.\ most straight-driving scenarios) and also requires tremendous labelling efforts, it is practically infeasible.
Similarly, \cite{bahari2022vehicle} proposes to augment the training dataset with challenging samples to fine-tune the models, which only alleviates the problem slightly.
As will be shown, removing the past motion history from the inputs is problematic as well since it contains vital clues for the prediction task.
\subsection{Frenet Frame Wrapper} 
We demonstrate how this generalisation problem can be effectively addressed by leveraging a scene representation that is well-known in ``classic'' planning approaches: the Frenet coordinate frame.
In contrast to previous works that have used the Frenet representation in their specific model architectures~\cite{gilles2022gohome,zhang2021map, song2022learning}, we present a general wrapper that is applicable to SotA models without the need to make changes to the architecture.
More specifically, we propose to first identify relevant lane centrelines in a traffic scene. 
Each of these centrelines defines a Frenet frame, for which the prediction model is queried. The model is trained to represent the prediction in the same Frenet frame, creating a strong inductive bias towards lane following.
Moreover, our wrapper allows us to query a prediction model multiple times depending on the number of relevant centrelines.
This promotes diversity and ensures that all relevant behaviours are captured in the set of predicted trajectories.

\subsection{Contribution}
In contrast to previous methods, we propose to use the Frenet representation in a wrapper compatible with SotA prediction models. As this is independent of a specific model architecture, it lets us analyse the regularising effect of the Frenet representation.
We summarise our main contributions to robust motion prediction in a Frenet frame as follows:
\begin{itemize}
    \setlength{\itemsep}{0pt}
    \item We propose a wrapper compatible with SotA prediction models, that leverages a Frenet representation of the scene.
    \item We conduct experiments with two different SotA prediction models and show that our wrapper reduces off-road predictions in challenging scenarios by a large margin and highlight the tradeoff between prediction accuracy and generalisation.
    \item We demonstrate how the additional benefit of a map-adaptive number of forecasted trajectories results in more diverse and accurate predictions.
\end{itemize}

\subsection{Structure}
The remainder of the paper is structured as follows.
Section~\ref{sec:related_work} gives an overview of relevant work on trajectory prediction in the Frenet frame as well as on generalisation and robustness in vehicle trajectory prediction.
Subsequently, in Section~\ref{sec:formulation} and~\ref{sec:method_description} we describe our method before introducing our experimental setup in Section~\ref{sec:experiments}.
Results are given in Section~\ref{sec:results}.
Finally, in Section~\ref{sec:conclusion} we draw conclusions and point out promising directions for future research.

%% file: chapters/2_related_work.tex
\section{Related work}
\label{sec:related_work}
\subsection{Trajectory Prediction in Frenet Frame}
The task of motion prediction is to forecast the future trajectory of a target vehicle (TV) within a traffic scene, given the scene context such as the past trajectory of the TV and surrounding road users as well as map information.
Commonly model inputs and output trajectories are represented in a Cartesian coordinate frame centred around the TV with the $x$-axis pointing forward and the $y$-axis pointing to the left \cite{gao2020vectornet,gilles2022gohome,liang2020learning,varadarajan2022multipath++,hallgarten2023prediction}.
At the same time the curvilinear coordinate system along a selected lane centreline, termed Frenet frame, is a natural choice to represent vehicle trajectories as it decomposes trajectories into their progress along a reference lane and the lateral displacement across it. 
Thus, it has been broadly adopted in planning \cite{pulver2021pilot, li2022human, kant1986toward, werling2010optimal, huang2022differentiable} and prediction based on heuristics \cite{yao2013lane, houenou2013vehicle}.
In data-driven prediction, however, only few works leverage this representation.
For instance, \cite{zhang2021map} represents the variable number of lanes in a traffic scene with a raster in the Frenet frame and subsequently models interaction between them using a GNN formulation. Finally, trajectories are decoded for each lane separately.
To generate a heatmap of potential trajectory endpoints, GOHOME~\cite{gilles2022gohome} first produces Frenet frame heatmaps for each relevant lane, projects them back to the Cartesian frame and superposes them. 
Similarly, PRIME~\cite{song2022learning} generates feasible trajectories in a Frenet Frame (cf. ~\cite{werling2010optimal}) and then utilises a learning-based evaluator to select a set of potential future trajectories.
Unlike previous methods, our approach is not specific to a particular model architecture.
Instead, we present a general wrapper that can be applied to a large variety of SotA prediction models without changing their architecture.
Hence we are able to study the tradeoff between prediction accuracy and generalisation that this representation induces.

\subsection{Robustness in Trajectory Prediction}
Many previous works studied the robustness of data-driven prediction models, e.g.\ by adapting the target agent~\cite{saadatnejad2021shared} or surrounding agents ~\cite{abeysirigoonawardena2019generating, althoff2018automatic, klischat2019generating} in an adversarial manner, with the goal of tricking the prediction model into forecasting an unrealistic trajectory.
However, it is non-trivial to define when a prediction model has been fooled in these augmented scenarios. 
For instance, if the TV swerves in its lane, predicting that it will drive into the opposite lane could well be reasonable.
Other works focus on generating new road topologies instead of augmenting surrounding agents.
For instance, \cite{saadatnejad2021shared, mi2021hdmapgen} utilize generative models to create novel maps.
However, they are not necessarily realistic, mainly due to potential artifacts.
In contrast, \cite{bahari2022vehicle} proposes to generate maps by perturbing existing scenes from a public benchmark \cite{chang2019argoverse}.
In particular, the road topology is augmented by adding turns in front of the TV, while the road behind the TV and its motion history are left unchanged.
The resulting maps are shown to resemble roads and intersections in the real world, hence it is confirmed that these perturbations are indeed realistic.
At the same time, the generated scenarios are challenging, as the model has to predict the TV to turn in, i.e.\ significantly deviate from the extrapolation of the past motion.
It is shown that 60\% of the benchmark scenes can be modified so that prediction methods fail.
In this work, we use this benchmarking methodology to evaluate our approach and show that integrating SotA prediction models in a Frenet wrapper reduces off-road predictions by over 90\%. 

%% file: chapters/3_notation.tex
\section{Notation and Definitions}
\label{sec:formulation}
Prediction models forecast a TV's future behaviour conditioned on the past trajectories of the TV and surrounding agents, as well as an HD Map describing the road topology.
Commonly, past trajectories are represented as a sequence of states $\mathbf{s}=\{s_{t}^{i}\}_{t=-t_h}^0$ for agent $i$ with $t_h$ the horizon of past timesteps.
The state vectors $s_t^i \in \mathbb{R}^m$ may include position, orientation, velocity, yaw rate, and acceleration.
The map is represented as a set of $l=1,..,L$ lanes given by a sequence of $n=1,..,N_l$ poses on their centrelines $\mathbf{q}=\{q_{n}^{l}\}_{n=1}^{N_l}$.
Furthermore, lane connections are described by an adjacency matrix.
A set of $k=1,.., K$ trajectories $t_k$ is predicted, and a probability $p(t_k)$ is assigned to each. 

%% file: chapters/4_method.tex
\section{Method description}
\label{sec:method_description}

\begin{figure*}
    \centering
    \includegraphics[width=\textwidth]{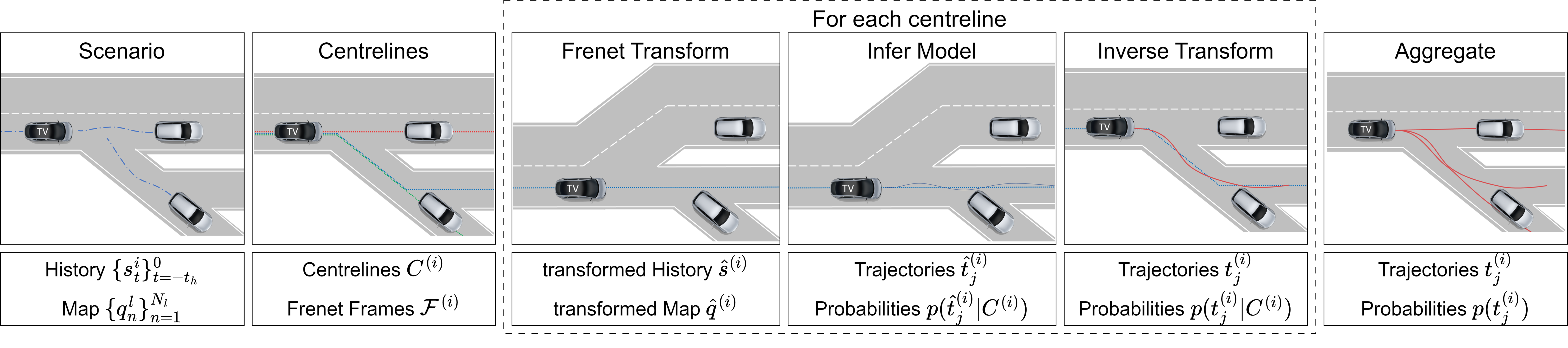}
    \caption{\textbf{Method overview} After identifying relevant lane centreline sequences for the TV, the scene is transformed to the corresponding Frenet frame of each centreline. Next, the prediction model is inferred and the output is projected back to the cartesian frame. Finally, predictions from all centrelines are aggregated. For simplicity, the figure shows $K=1$ predictions per centreline.
    }
    \label{fig:method_overview}
\end{figure*}
We propose to integrate SotA prediction models in a Frenet frame wrapper. 
We defer a summary of the necessary foundations to the supplementary material.
An overview of our method is shown in Fig.~\ref{fig:method_overview}.
It consists of the following steps:
First, assign the TV to a lane centreline and identify all relevant $N$ sequences of lane centrelines.
Second, iterate over the centrelines and transform the full scene representation into the Frenet frame of that centreline.
Then query the prediction model to obtain $K$ trajectories for each of the $N$ centrelines, totalling $K\cdot N$ trajectories $t_k^{(n)}$ each assigned an (unnormed) probability $p_k^{(n)}$.
To form the final set of output trajectories, suppress near-duplicate trajectories and select a set of $\hat{K}$  trajectories based on their probabilities.

\subsection{Identification of Relevant Centrelines}
We infer the prediction model for all lane centerline sequences the TV could traverse.
For our experiments, we use the closest lane with an orientation offset below $\frac{\pi}{4}$ as the current lane and extend it to a length of $110$ metres.
At intersections and forks, we duplicate the centreline and follow each successor individually. 
Note that this may result in identical centreline sequences up to a very distant point, resulting in nearly identical inputs for the prediction model. 
However, we decide to deal with these near-duplicates at the trajectory level, rather than at the centreline level.
We denote the sequences of centrelines $C^{(i)} \in \mathbb{R}^{m\times 2}, i=1,..N$ and the respective Frenet frame by $\mathcal{F}^{(i)}$
where $m$ is the number of 2D, i.e., $(x,y)$, poses along $C^{(i)}$.

\subsection{Frenet Frame Transformation}
We transform $\mathbf{s}$ and $\mathbf{q}$ into each Frenet frame $\mathcal{F}^{(i)}$. We chose the origin of $\mathcal{F}^{(i)}$ such that the TV's current position is at $s=0$. Coordinate frame-independent attributes (e.g., speed) remain unchanged. Additionally, we replace the yaw angle in $\mathbf{q}$ with the curvature.
We give a thorough description of our transformation algorithm in the supplementary material, where we also discuss corner cases in which the transformation is ambiguous. Besides, we discuss the computational burden that this transformation induces.

\subsection{Inferring the Prediction Model}
We infer the prediction model using the inputs $\mathbf{\hat{s}^{(i)}}$ and $\mathbf{\hat{q}^{(i)}}$ for each Frenet frame $\mathcal{F}^{(i)}$.
The model's architecture is left unchanged.
Predicted trajectories $t_j^{(i)}$ are represented in $\mathcal{F}^{(i)}$ and probabilities $p(t_j^{(i)}|C^{(i)})$ are conditioned on the centreline $C^{(i)}$.
At inference time, the trajectories are projected back to the Cartesian frame.
We can infer the predictions for a varying number of identified $N$ centrelines with a single batched forward pass to reduce computational overhead and obtain a variable number of $K\cdot N$ output trajectories.

\subsection{Output Aggregation}
We believe that a variable number of modes is not bad per se, as the number of potential future behaviours strongly depends on the traffic scene.
However, to comply with public benchmarks, we reduce the number of output trajectories to a fixed number $\hat{K}$. 
We estimate the prior $p(C^{(i)})$ with a simplistic MLP model that predicts a score for each centreline conditioned on the target vehicle's past motion history and a set of points along the centreline, both represented in the original Cartesian frame.
A subsequent softmax layer over all centrelines yields the probability of each lane $p(C^{(i)})$, which is used to obtain the probabilities of the trajectories $p(t_i)$ by marginalising the predicted conditioned probabilities $p(t_i|C^{(i)})$.
Similar to~\cite{gilles2021home}, we then greedily pick a set of $N$ trajectories, starting from the one with the highest probability and suppressing all trajectories whose endpoint is located within a radius of $1.0$ metres.
We give a thorough description of the lane scoring model and ablations in the appendix, where we also report experiments with alternative methods such as assuming a uniform prior or using $K$-means clustering as in~\cite{deo2022multimodal}.

\subsection{Learning}
We want to teach the model a strong inductive bias towards following a given lane.
Since only a single ground-truth trajectory is available during training, we determine the corresponding centreline $C^{*}$ based on minimal average displacement.
Subsequently, we infer the model only with respect to the respective Frenet frame $\mathcal{F}^{*}$.
We use the inputs $\mathbf{\hat{s}^{*}}$ and $\mathbf{\hat{q}^{*}}$ represented in $\mathcal{F}^{*}$ to obtain $K$ trajectories $t_j^{*}$ and probabilities $p(t_j^{*}|C^{*})$.
After transforming the ground truth to $\mathcal{F}^{*}$, we apply the same loss function as for the model trained on cartesian data.

%% file: chapters/5_evaluation.tex
\section{Evaluation}
\label{sec:experiments}

\subsection{Benchmark}
\begin{figure*}
    \centering
    \begin{subfigure}[b]{0.20\textwidth}
        \centering
        \includegraphics[width=\textwidth]{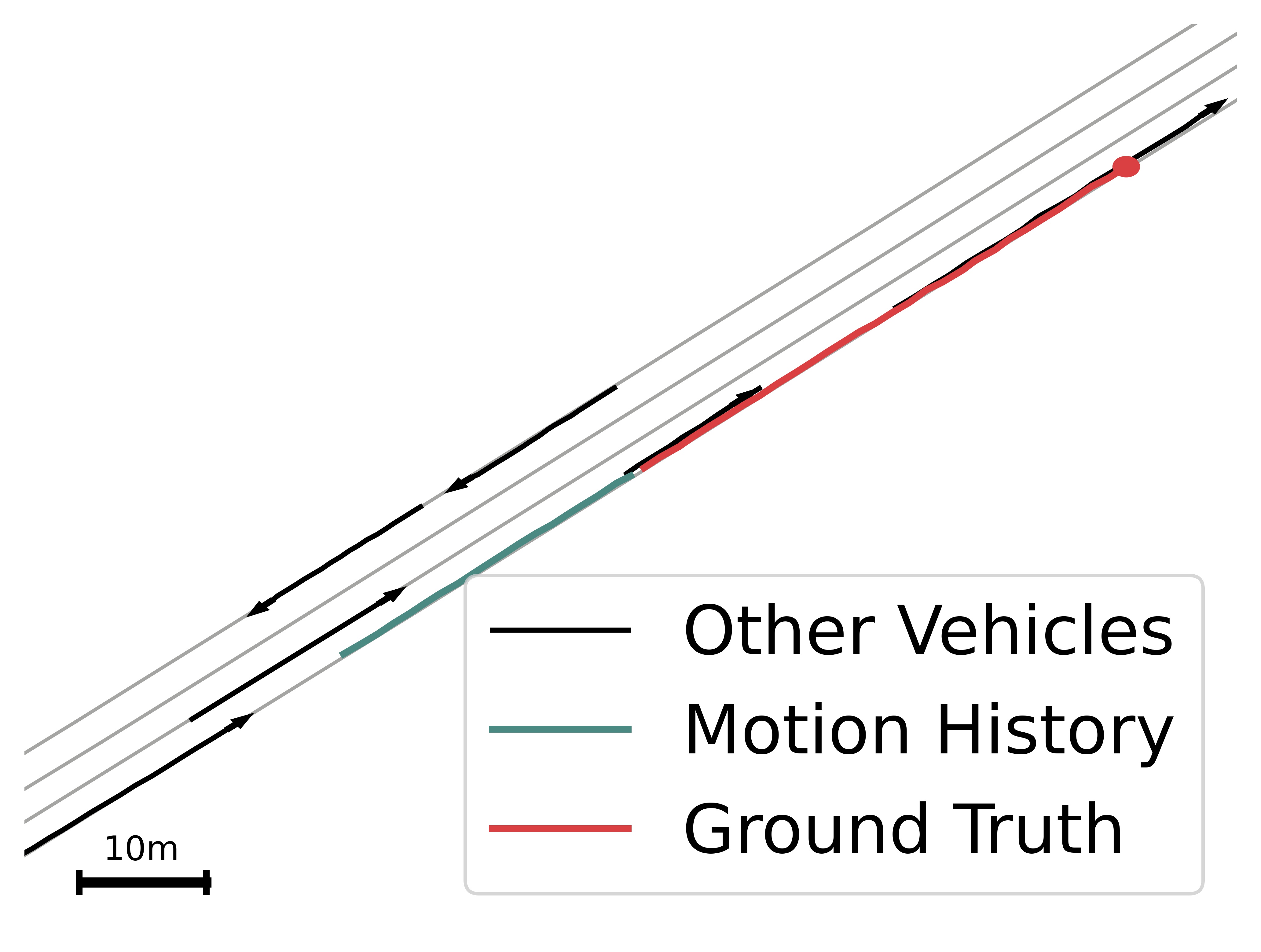}
        \caption{Original Scene}
    \end{subfigure}%
    \hfill
    \begin{subfigure}[b]{0.20\textwidth}
        \centering
        \includegraphics[width=\textwidth]{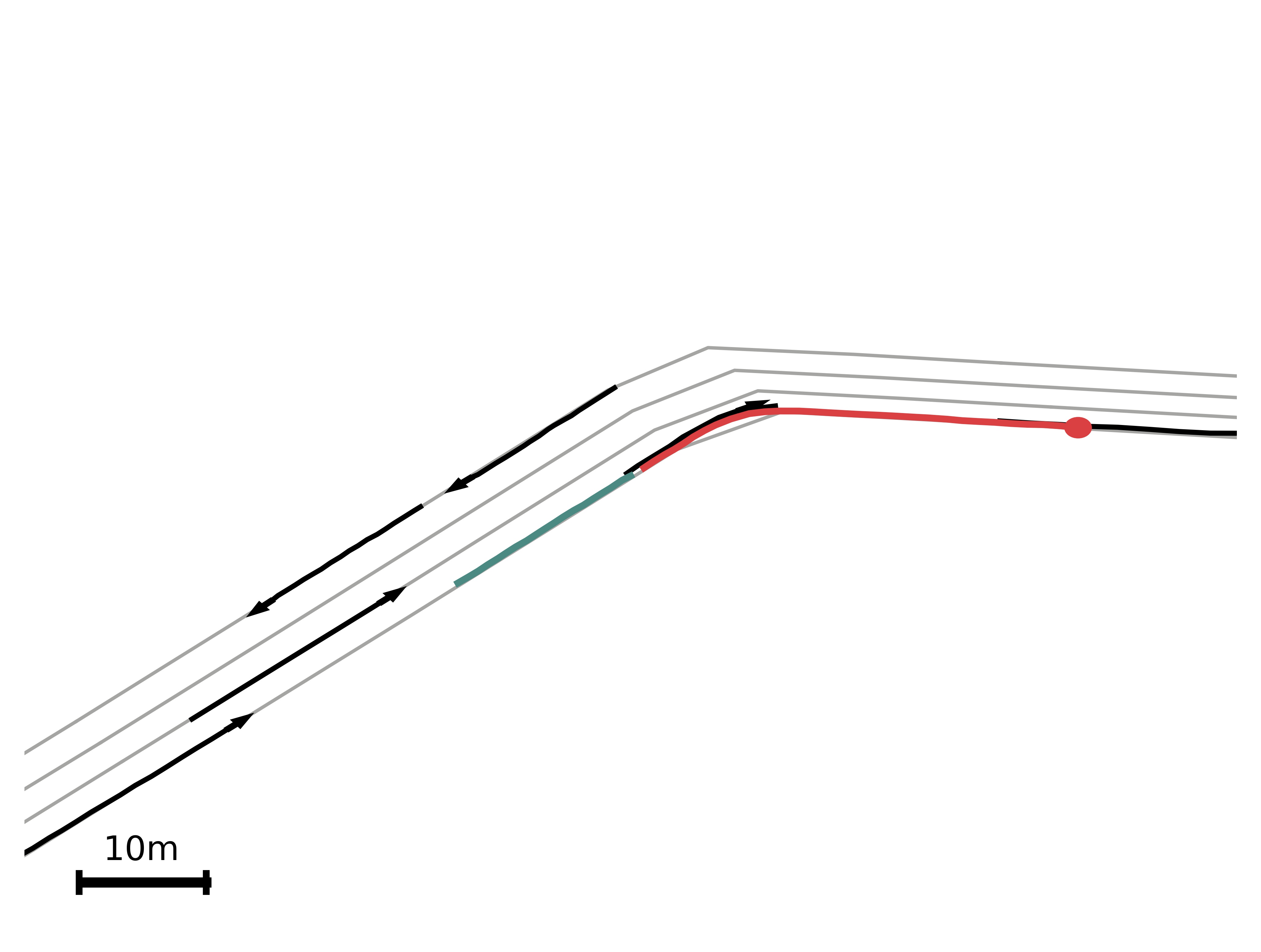}
        \caption{Smooth Turn}
    \end{subfigure}%
    \hfill
    \begin{subfigure}[b]{0.20\textwidth}
        \centering
        \includegraphics[width=\textwidth]{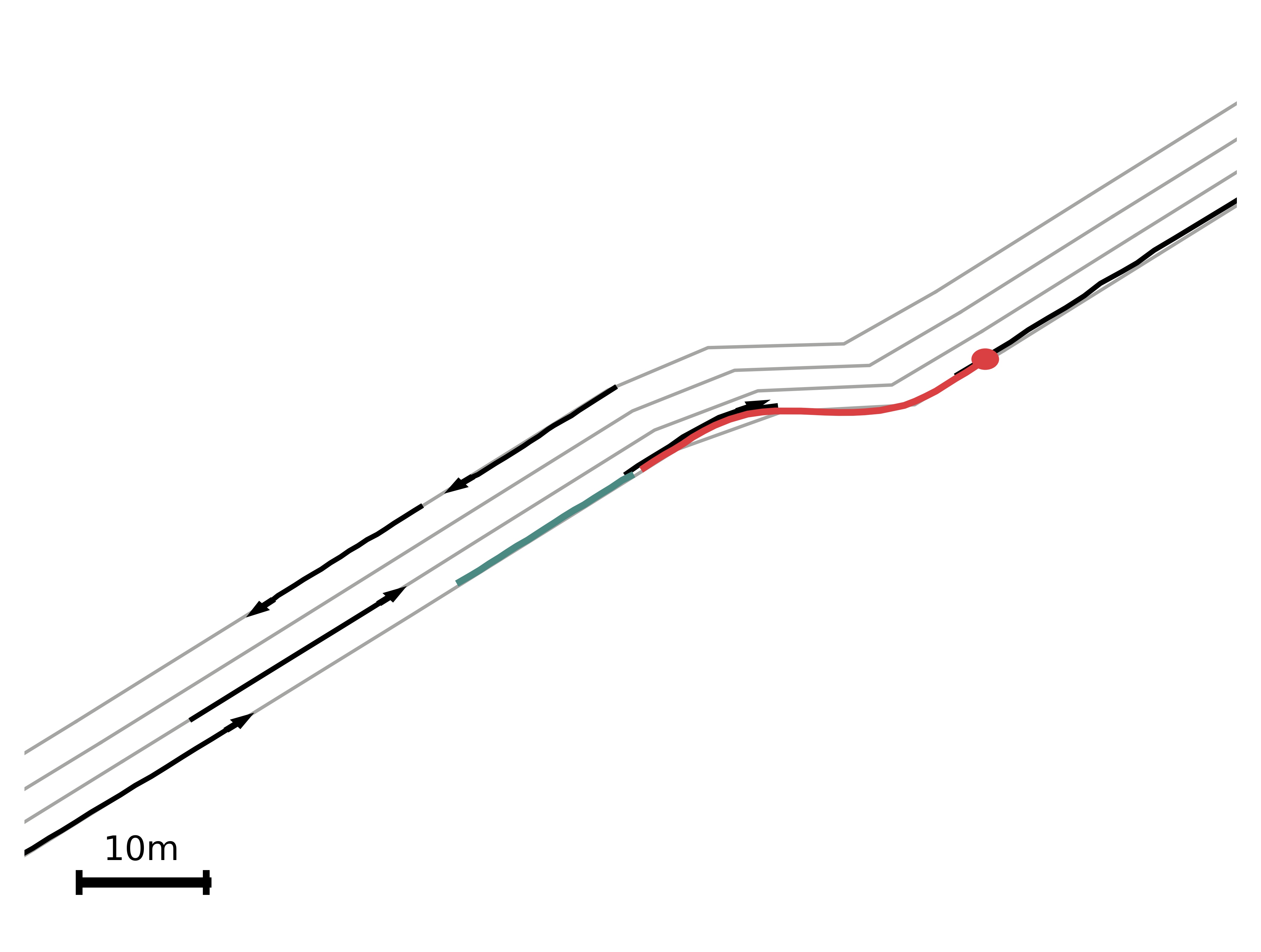}
        \caption{Double Turn}
    \end{subfigure}%
    \hfill
    \begin{subfigure}[b]{0.20\textwidth}
        \centering
        \includegraphics[width=\textwidth]{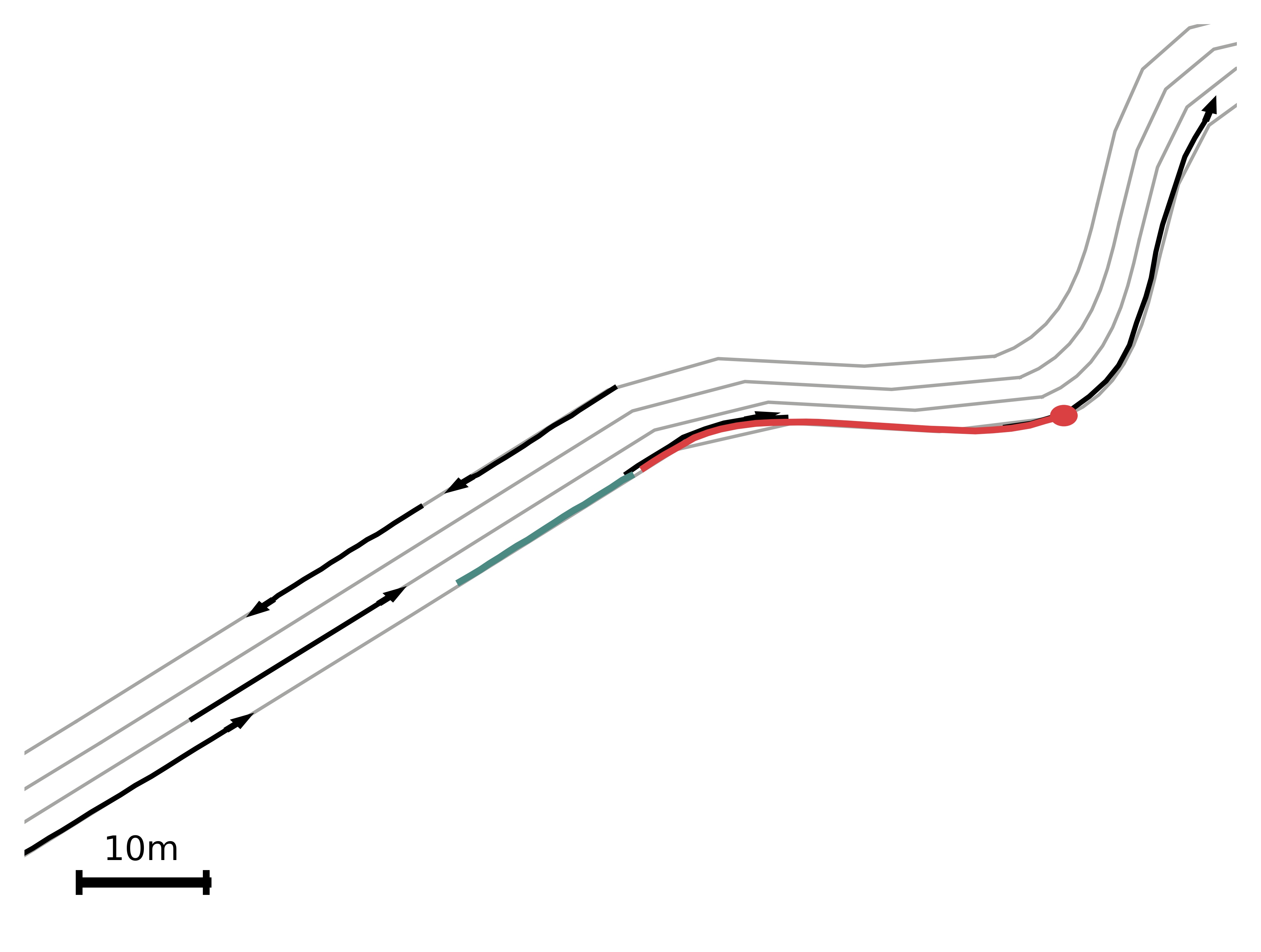}
        \caption{Ripple Road}
    \end{subfigure}
    \caption{\textbf{scene-attack perturbations.} Three different transformations are applied to the original scene. All of them create a turn at a fixed distance ahead of the target vehicle. Observe how the motion history and the ground truth are slowed down to account for the large curvature.}
    \label{fig:scene_attack}
\end{figure*}

We evaluate our method on the Argoverse~\cite{chang2019argoverse} dataset. Collected in Pittsburgh and Miami, this large-scale dataset comprises 205,942 scenarios for training and 39,472 for testing. Each sample consists of 2 seconds of history with the goal of predicting the next 3 seconds of the TV's future motion.
We leverage the scene-attack benchmark~\cite{bahari2022vehicle} to generate new challenging test scenarios and conduct experiments on the generalisation ability of our method.
This benchmark generates new scenes conditioned on existing scenes by shifting ``scene points'' according to different transforms. 
Scene points include the observed past trajectories of all agents, the ground truth future of the TV, and points describing the lane centrelines. 
The transform functions only modify scene points after a predefined distance $b$ ahead of the TV, leaving the road behind and immediately ahead of it unchanged.
To make sure the generated scenario is still feasible, velocities may be reduced such that the curve ahead can be passed safely at the current speed, i.e.\ without exceeding a lateral acceleration of $0.7g$.
Fig.~\ref{fig:scene_attack} shows the three attack functions we use to perturb an original scene:
smooth turn, double turn, and ripple road.
Each attack function is applied in both directions (left and right), and we report the worse result. 

\subsection{Metrics}
We use the following metrics to evaluate our results.
Off-Road Probability (ORP) is the cumulative probability of all predicted trajectories with at least one waypoint off-road.
To quantify the diversity of the predicted set of trajectories, we use Mean Inter Endpoint Distance (MIED), the average distance of all predicted endpoints to the mean endpoint.
We further report the commonly used metrics for trajectory prediction, namely minimum Average / Final Displacement Error (minADE / minFDE).
Additionally, we use single-trajectory Miss Rate (MR1) to evaluate the ratio of scenarios in which the endpoint of the trajectory with the highest probability is more than $2.0$ metres away from the ground truth.
Since we do not have a ground truth future trajectory for perturbed scenes, we generate pseudo-labels, by transforming the future trajectory in the same way as the scene points.
Alike to the motion history, the future motion may be scaled to a lower speed to ensure physical feasibility.

\subsection{Models}
We test our wrapper approach with two SotA prediction models: \lanegcn and \multipath.
Code for \lanegcn is publicly available and was among the best contestants in the Argoverse Forecasting Challenge 2020. 
Besides, we use the faithful reimplementation of \multipath given in \cite{konev2022mpa}. 
We compare the prediction performance of the original models with those integrated into our Frenet wrapper and denote the latter with the suffix \textit{SD} (as Frenet coordinates are commonly denoted as $(s,d)$).
\lanegcn is trained for 36 epochs with a batch-size of 32. 
The learning rate is decayed from $10^{-3}$ to $10^{-4}$ after 32 epochs.
\multipath is trained for 100 epochs with a batch-size of 128, an initial learning rate of $10^{-4}$ that is decayed by 0.5 every 20 epochs. 

\subsection{Baselines}
Off-road predictions indicate that the predictor does not consider the map information sufficiently.
In particular, we notice that the predicted trajectories of the Cartesian models hardly change with the perturbations (see Fig.~\ref{fig:qualitative_results}).
Hence, we speculate that the models tend to extrapolate the motion history.
We compare our method to two simple baselines that address this problem.
First, we add a training regularisation that prevents the model from only extrapolating the history by applying dropouts to the TV's motion history in 50\% of the training samples.
We denote this with the suffix \textit{DO}.
Second, we remove the motion history entirely from the model input at training and inference time (suffix \textit{NH}).
As augmenting the training dataset with perturbed scenarios has been shown to only alleviate the off-road prediction problem slightly~\cite{bahari2022vehicle}, we train neither the Cartesian nor the Frenet models with data augmentation.
To emphasise that our Frenet frame wrapper surpasses a simple lane following baseline, we additionally evaluate a simple heuristic model: predicting trajectories with fixed accelerations $a\in\{-4,2,0,2,4,a_t\}\frac{m}{s^2}$, where $a_t$ is the current acceleration.
This baseline is also evaluated in the Cartesian frame (\textit{CA}) as well as in the Frenet frame (\textit{CA-SD}).

%% file: chapters/6_results.tex
\section{Results}
\label{sec:results}
\begin{table}
    \small
    \centering
    \adjustbox{max width=\columnwidth}{
        \begin{tabular}{ lllllll }
        \toprule
        \textbf{Model}  & \textbf{$\hat{K}$}& \textbf{minADE}~$\downarrow$& \textbf{minFDE}~$\downarrow$& \textbf{ORP}~$\downarrow$& \textbf{MR1}~$\downarrow$& \textbf{MIED}~$\uparrow$\\
        & &[m]&[m]&[\%]&[\%]&[m]\\
        \midrule
        \multipath      & $6$          & $0.772$ & $1.399$ &  $1.7$ & $56.7$& $2.908$\\
        MultiPath++SD   & $6$          & $0.837$ & $1.470$ &  $1.2$ & $59.8$& $3.216$\\
        MultiPath++SD   & $N\cdot 6$   & $0.767$ & $1.274$ &  $1.4$ & $59.8$& $3.473$\\
        \lanegcn        & $6$          & $0.645$ & $1.076$ &  $0.7$ & $48.7$& $3.800$\\
        LaneGCN-SD      & $6$          & $0.737$ & $1.233$ &  $1.1$ & $54.1$& $4.261$\\
        LaneGCN-SD      & $N\cdot 6$   & $0.704$ & $1.138$ &  $1.3$ & $54.1$& $4.678$\\
        \midrule
        MultiPath++DO   & $6$          & $0.784$ & $1.425$ &  $1.6$ & $56.8$& $2.793$\\
        MultiPath++NH   & $6$          & $1.137$ & $2.028$ &  $1.8$ & $75.5$& $4.813$\\
        LaneGCN-DO      & $6$          & $0.659$ & $1.114$ &  $0.7$ & $52.0$& $3.889$\\
        LaneGCN-NH      & $6$          & $0.983$ & $1.665$ &  $1.0$ & $59.7$& $5.729$\\
        CA-SD           & $N\cdot 6$   & $2.410$ & $3.745$ &  $0.1$ & $-$& $12.180$\\
        CA              & $6$          & $2.659$ & $4.669$ & $14.5$ & $-$& $12.050$\\
        \bottomrule
        \end{tabular}
    }
    \caption{\textbf{Results on original scenes}}
    \label{tab:results}
\end{table}
In the following, we present our results on the original scenes from the Argoverse dataset and the scene-attack benchmark.
Based on these results, we want to answer the following questions:
\begin{itemize}
    \item Does the Frenet wrapper result in predictions that are competitive with Cartesian models?
    \item Do models trained with our Frenet wrapper generalise beyond simple lane following? 
    \item Can we use the Frenet representation to increase the robustness of SotA predictors?
\end{itemize}
\subsection{Comparison on Original Scenes}
We report our results on the unperturbed Argoverse benchmark in Tab.~\ref{tab:results}.
We find that, the \textit{SD} models with $\hat{K}=6$ have higher displacement errors than the corresponding Cartesian models.
At the same time, the off-road probability is of equally small magnitude and the predictions are more diverse.
We observe that the increase in ORP of $0.4$ percent of the LaneGCN-SD model results from scenarios where it fails to accurately follow a different lane than the reference lane.
Note that an ORP of 1 percent indicates that the off-road trajectories amount to a probability of only $0.01$ on average.

\begin{figure*}
    \centering
    \adjustbox{max width=\textwidth}{
    \large
    \begin{tabular}{>{\centering\arraybackslash}p{0.22\textwidth}>{\centering\arraybackslash}p{0.22\textwidth}>{\centering\arraybackslash}p{0.22\textwidth}>{\centering\arraybackslash}p{0.22\textwidth}}
        \multipath& Multipath++SD& \lanegcn& LaneGCN-SD\\
    \end{tabular}
    }
    \includegraphics[width=\textwidth]{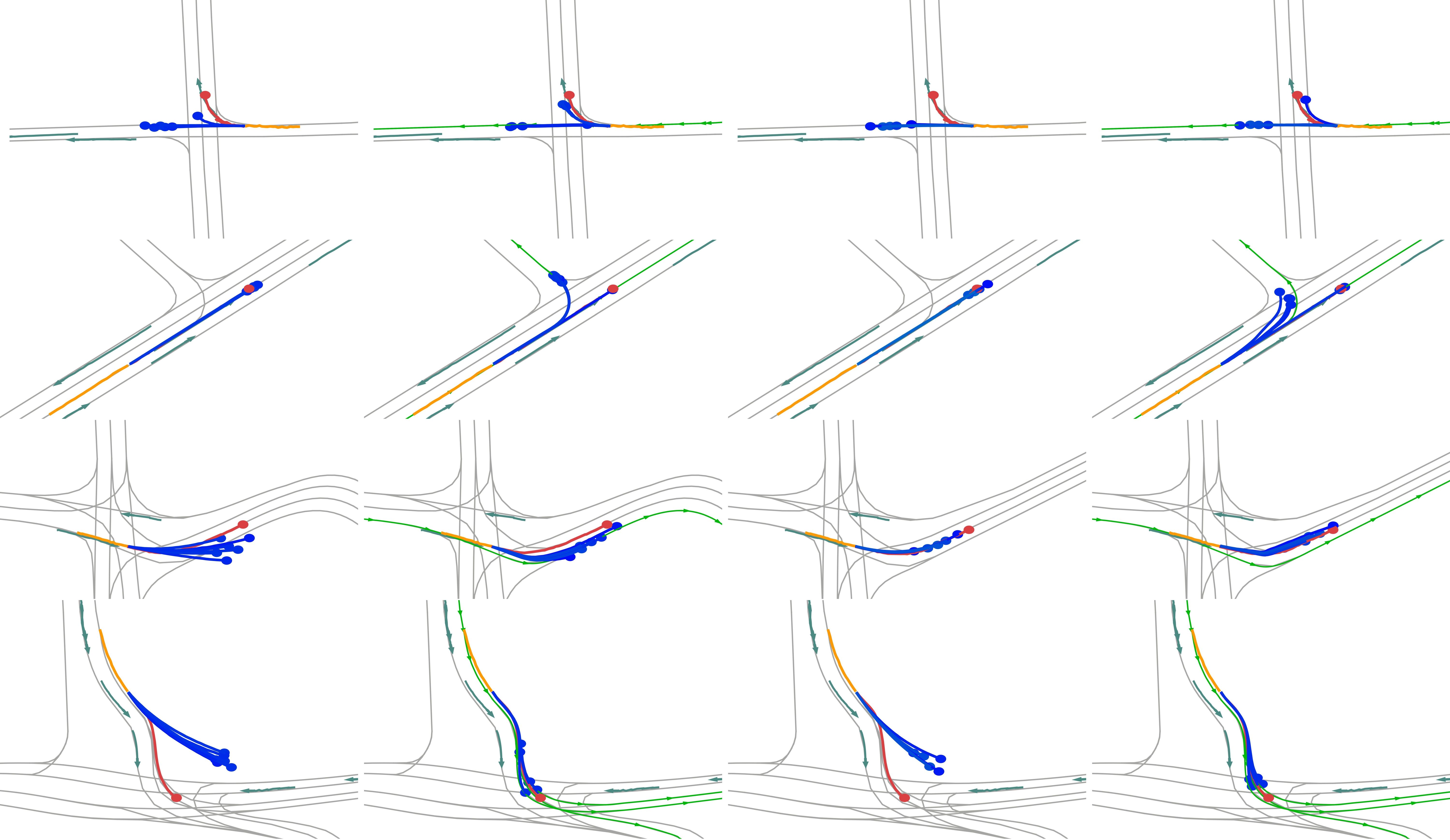}
    \caption{\textbf{Qualitative Results} From top to bottom two original and two perturbed scenes are shown. Centrelines used by the \textit{SD} models are shown in green, state history in yellow. Predicted trajectories and ground truth are shown in blue and red, respectively. In the top example, the TV has just passed the branching point so that only the straight centreline is detected. The \textit{SD} models are still able to predict the right-turn mode, significantly off the reference centreline. The second row demonstrates the increased diversity of \textit{SD} model predictions. The third row shows deviations from the reference line. In the bottom example, the \textit{SD} models adapt to a difficult perturbation, while the original models predict only off-road trajectories (Best viewed in colour).}
    \label{fig:qualitative_results}
\end{figure*}

Our qualitative results (see Fig.~\ref{fig:qualitative_results}) also show that the \textit{SD} models do not simply follow the lane centrelines.
On the contrary, we observe how they adapt to corner cases where a branching centreline is not detected because the TV has just moved past the branching point, by following another lane than the reference lane. 
Additional qualitative results, which demonstrate lane changes and illegal manoeuvres can be found in the supplementary material.

Unsurprisingly, the models trained and evaluated without motion history (\textit{-NH}) perform far worse than the models that had access to past observations. 
In contrast, the models trained with dropouts on the motion history (\textit{-DO}) achieve competitive performance compared to the models that had access to it at all times.
However, as we will demonstrate later, this does not improve generalisation capabilities to perturbed scenes.
As expected, the Constant-Acceleration model (CA) is a poor contestant with the highest displacement errors and off-road probabilities.
Applying our Frenet wrapper to this simple heuristic significantly reduces the displacement errors.
At the same time, the off-road probability (ORP) drops, as expected, to almost zero.
The small remaining ORP is caused by samples where the predicted trajectory exceeds the map limits.

\begin{table*}
    \centering
    \adjustbox{width=\the\textwidth}{
    \begin{tabular}{llllllllllllll}
    \toprule
        \multirow{2}{*}{Model}&\multirow{2}{*}{$\hat{K}$}& &\multicolumn{3}{c}{Single Turn} & &  \multicolumn{3}{c}{Double Turn}& &  \multicolumn{3}{c}{Ripple Road} \\
        \cline{4-6}\cline{8-10}\cline{12-14}
        &&&minADE& ORP& MR1&& minADE& ORP& MR1&& minADE& ORP& MR1\\
    \midrule
    \multipath      & $6$          && $3.139$ & $65.5$ & $96.5$ && $2.574$& $67.1$ & $96.3$ && $3.549$& $72.9$ & $97.9$\\
    MultiPath++SD   & $6$          && $1.388$ &  $3.6$ & $88.0$ && $1.132$& $3.7$ & $78.5$ && $1.331$&  $3.5$ & $85.2$\\
    MultiPath++SD   & $N\cdot 6$   && $1.265$ &  $3.7$ & $88.2$ && $1.018$&  $3.8$ & $78.8$ && $1.265$&  $3.7$ & $85.2$\\
    \lanegcn        & $6$          && $2.063$ & $38.2$ & $93.9$ && $2.372$& $63.9$ & $94.4$ && $3.094$& $61.2$ & $96.2$\\
    LaneGCN-SD      & $6$          && $1.209$ &  $3.1$ & $85.6$ && $0.987$&  $3.5$ & $70.3$ && $1.144$&  $2.9$ & $81.7$\\
    LaneGCN-SD      & $N \cdot 6$  && $1.146$ &  $3.3$ & $85.8$ && $0.940$&  $3.7$ & $70.7$ && $1.075$&  $3.0$ & $81.7$\\
    \midrule
    MultiPath++DO   & $6$          && $3.205$ & $97.8$ & $56.8$ && $2.594$& $69.7$ & $97.7$ && $3.457$& $74.5$ & $98.9$\\
    MultiPath++NH   & $6$          && $5.653$ & $98.7$ & $75.5$ && $3.459$& $73.8$ & $97.4$ && $4.264$& $78.1$ & $98.7$\\
    LaneGCN-DO      & $6$          && $2.209$ & $45.7$ & $95.2$ && $2.312$& $59.6$ & $95.1$ && $2.937$& $57.2$ & $96.6$\\
    LaneGCN-NH      & $6$          && $2.652$ & $54.7$ & $97.1$ && $3.150$& $68.6$ & $97.0$ && $3.705$& $67.8$ & $98.1$\\
    CA-SD           & $N \cdot 6$  && $2.209$ & $0.5$  & $-$    && $2.175$&  $1.1$ & $-$    && $2.251$&  $0.0$ & $-$\\
    CA              & $6$          && $4.748$ & $58.2$ & $-$    && $4.071$& $57.6$ & $-$    && $5.208$& $61.9$ & $-$\\
    \bottomrule
    \end{tabular}
    }
    \caption{\textbf{Results on perturbed scenes}}
    \label{tab:results_perturbed}
\end{table*}

\subsection{Comparison on Perturbed Scenes} 
In the following, we analyse our results on the scene-attack benchmark (see Tab.~\ref{tab:results_perturbed}) to evaluate the robustness of the models under map perturbations.
For the Cartesian models, we observe that displacement errors and ORP increase drastically when perturbing the scenes.
We observe an ORP increase from 0.7\% to 63.9\% and from 1.7\% to 67.1\% in the Double Turn perturbation for \lanegcn and \multipath respectively. 
This is consistent with what has been observed by~\cite{bahari2022vehicle}.
In contrast, \textit{SD} models are much more robust, with ORP only increasing by 2.5 percentage points for both models.
Note that the perturbed scenarios are more challenging than the unperturbed ones, e.g.\ there are no straight driving scenarios, which explains the moderate increase in ORP.
Additionally, the pseudo-ground truth used to evaluate the displacement errors causes uncertainty in the calculation of these metrics.
Our results also show that neither Dropouts in training (\textit{DO}) nor removing the motion history entirely from the features (\textit{NH}) make the models more robust. 
Overall, the \textit{SD} models outperform all the baselines by a large margin, resulting in a new State-of-the-Art on the scene-attack benchmark.

\subsection{Discussion}
The qualitative results shown in Fig.~\ref{fig:qualitative_results} demonstrate how the SD models generalise seamlessly to the perturbed scenes, whereas the Cartesian models fail to adapt.
Furthermore, we find that the remaining off-road predictions of the \textit{SD} models are mostly caused by corner cases in the identification of relevant centrelines.
For instance, in cases where the TV is positioned directly behind a lane branching point, assigning it to a single lane results in not detecting the other lane as a relevant centerline.
We find that the model partially compensates for this, by also following other lanes than the reference lane of the coordinate frame.
However, we observe cases in which some waypoints of the resulting trajectories are slightly off-road.
Nonetheless, the results clearly show that the models do not blindly follow the centreline of the Frenet frame.
On the contrary, as can be seen in the qualitative results, the models are able to deviate significantly from it, indicating a thorough understanding of the scene.

%% file: chapters/7_conclusion.tex
\section{Conclusion}
\label{sec:conclusion}
In this work, we presented a general Frenet transform wrapper compatible with SotA prediction models and demonstrated its effectiveness on two different SotA prediction models.
We found that representing the input and output of prediction models in the Frenet frame improves generalisation to challenging scenarios by a large margin. 
Hence, we were able to reduce predicted off-road probability by over 90\% on the scene-attack benchmark.
Moreover, our wrapper comes with the advantage of more diverse predictions, increasing MIED (a measure for the diversity of predicted trajectories) by 20\%, at the cost of slight performance decrease on original scenarios.

%% file: chapters/8_limitations.tex
\section{Limitations}
In the case of a fixed number of predicted trajectories, the robustness of our method comes at the cost of slightly reduced prediction accuracy and inference runtime.
Though, our method can infer predictions for multiple centrelines with a single batched forward pass and we provide an efficient implementation for the coordinate transformation.
Moreover, our results indicate that a better estimation of the prior over the relevant lane sequences can enhance the predicted mode probabilities.

%% file: chapters/appendix.tex
\begin{table*}
    \small
    \centering
    \adjustbox{max width=\the\textwidth}{
        \begin{tabular}{ llllllll }
        \toprule
        \textbf{Model} & \textbf{prior/mode selection} & \textbf{$\hat{K}$}& \textbf{minADE [m]}~$\downarrow$& \textbf{minFDE [m]}~$\downarrow$& \textbf{ORP [\%]}~$\downarrow$& \textbf{MR1 [\%]}~$\downarrow$& \textbf{MIED [m]}~$\uparrow$\\
        \midrule
        \multipath      & -             & $6$          & $0.772$ & $1.399$ &  $1.7$ & $56.7$& $2.908$\\
        Multipath++SD*  & privileged   & $6$          & $0.808$ & $1.390$ &  \textbf{1.2} & \textbf{59.3}& $3.098$\\
        MultiPath++SD-UP& uniform / all & $N\cdot 6$   & \textbf{0.767} & \textbf{1.274} &  $1.4$ & $61.7$& \textbf{3.473}\\
        \midrule
        Multipath++SD-GS& greedy-sampling& $6$          & \textbf{0.837} & \textbf{1.470} & \textbf{1.2} & \textbf{59.8}& $3.216$\\
        Multipath++SD-LS& lane-scoring  & $6$          & $0.915$ & $1.688$ &  $1.5$ & \textbf{59.8}& $3.019$\\
        Multipath++SD-KM& clustering    & $6$          & $0.833$ & \textbf{1.470} &  $1.3$ & $65.3$& \textbf{3.518}\\
        MultiPath++SD-UP& uniform       & $6$          & $0.936$ & $1.742$ &  $1.6$ & $61.7$& $2.942$\\
        \midrule
        \lanegcn        & -             & $6$          &  \textbf{0.645} &  \textbf{1.076} &   \textbf{0.7} &  \textbf{48.7}& $3.800$\\
        LaneGCN-SD*     & privileged    & $6$          & $0.716$ & $1.173$ &  $1.1$ & $53.6$& $4.347$\\
        LaneGCN-SD-UP   & uniform / all  & $N\cdot 6$   & $0.704$ & $1.138$ &  $1.3$ & $55.4$&  \textbf{4.678}\\
        \midrule
        LaneGCN-SD-GS   & greedy-sampling& $6$          &  \textbf{0.737} &  \textbf{1.233} &  $1.1$ &  \textbf{54.1}& $4.261$\\
        LaneGCN-SD-LS   & lane-scoring  & $6$          & $0.760$ & $1.292$ &   \textbf{1.0} &  \textbf{54.1}& $4.016$\\
        LaneGCN-SD-KM   & clustering    & $6$          & $0.759$ & $1.315$ &  $1.4$ & $65.1$&  \textbf{4.703}\\
        LaneGCN-SD-UP   & uniform       & $6$          & $0.777$ & $1.335$ &  $1.3$ & $55.4$& $4.091$\\
        \bottomrule
        \end{tabular}
    }
    \caption{\textbf{Ablations on centerline aggregation on original scenes}}
    \label{tab:appendix_results}
\end{table*}
\FloatBarrier
\appendix
In Tab.~\ref{tab:appendix_results}, we report results for an extensive set of methods to aggregate predictions conditioned on various centerliens. Models with greedy sampling, i.e., LaneGCN-SD-GS and Multipath++SD-GS are the methods used in our main paper.
\subsection{Privliedged Prior} Models denoted with an asterisk $^*$ only predict conditioned on the ground truth future centreline, i.e.\ they have perfect centreline scoring, and results serve as an upper bound.
We find that this significantly reduces the gap between the models with $\hat{K}=N\cdot6$ and the ones with $\hat{K}=6$. Moreover, Multipath++SD$^*$ even outperforms the Cartesian \multipath model.
This emphasizes that an accurate estimation of the prior over the lanes significantly improves prediction performance on the original scenes. 

\subsection{Uniform Prior}
In contrast to the privileged prior models, models denoted with $UP$ assume a uniform prior over the centrelines $C^{(i)}$, i.e.\ $p(C^{(i)}) = \mathcal{U}$.
Then, we pick the $\hat{K}$ trajectories with the highest probabilities $p(t_i)$, which are obtained by marginalising the predicted conditioned probabilities $p(t_i|C^{(i)})$.
We use the same uniform prior assumption in order to evaluate the miss-rate when selecting a variable number of trajectories $\hat{K}=N\cdot 6$.

\subsection{Lane Scoring Model}
We employ a simple feedforward model that predicts a score for each centreline in the scene. 
A subsequent softmax layer across all centrelines in the scene yields the probability for each lane.
The model inputs comprise the target vehicle's past trajectory as well as a sequence of poses along the centreline, given in Cartesian coordinates w.r.t.\ the current pose.
Both are independently processed by a single linear layer and then concatenated to form a feature vector.
Afterwards, an MLP with two hidden layers, each comprising 512 neurons, regresses the score for the given centerline.
The lane scoring model is trained for ten epochs at a learning rate of $1e-4$ with a batch size of 128. 
We use the cross-entropy loss as training objective.
We denote the models that are evaluated with the lane scoring prior with the suffix $LS$.

\subsection{Clustering}
As an alternative to estimating the prior, we employ a $K$-means clustering of the $N\cdot K$ trajectories to obtain a set of $K$ output trajectories~\cite{deo2022multimodal}. The estimated probability of each trajectory is obtained from the inverse rank of the cluster. We denote this ablation with the suffix $KM$.

\subsection{Greedy Sampling}
Similar to~\cite{gilles2021home}, the set of $K$ output trajectories can be selected greedily.
This extends the idea of the lane scoring model with a non-maximum suppression.
First, the lane scoring model is inferred to obtain the estimated probability of each trajectory.
Next, a greedy algorithm selects the most probable trajectory and removes all trajectories from the candidate set whose endpoint is within a radius of 1.0m of the selected one.
This is repeated until $K=6$ trajectories are selected.
We denote this model with the suffix $GS$.

\subsection{Results}
We find that selecting a set of $\hat{K}=6$ trajectories with a uniform prior results ($-UP$) in the highest displacement errors. As the ablation that leverages a privileged prior ($^*$) performs substantially better, we conclude that the impaired performance results from a suboptimal selection of the trajectories.
Consequently, leveraging the lane-scoring model ($LS$) to estimate the prior over the lanes results in decreased displacement errors.
Moreover, the miss-rate closely approaches the level of privileged prior ablation.
Combining this with non-maxima-suppression, as it's done by our greedy sampling method ($GS$), results in the lowest displacement errors among the $SD$ models while maintaining the low miss rate. 